\documentclass[conference]{IEEEtran}
\IEEEoverridecommandlockouts
\usepackage{cite}
\usepackage{amsmath,amssymb,amsfonts}
\usepackage{algorithmic}
\usepackage{multirow}
\usepackage[subrefformat=parens]{subcaption}
\usepackage{comment}
\usepackage[dvipdfmx]{graphicx}
\usepackage{textcomp}
\usepackage{url}
\usepackage{xcolor}
\usepackage{comment}
\usepackage{flushend}
\def\BibTeX{{\rm B\kern-.05em{\sc i\kern-.025em b}\kern-.08em
    T\kern-.1667em\lower.7ex\hbox{E}\kern-.125emX}}
\renewcommand\IEEEkeywordsname{Keywords}

\DeclareMathOperator*{\argmax}{arg\,max}

\makeatletter
\def\ps@IEEEtitlepagestyle{%
  \def\@oddfoot{\mycopyrightnotice}%
  \def\@evenfoot{}%
}
\def\mycopyrightnotice{%
  \begin{minipage}{\textwidth}
  \centering \scriptsize
  Copyright~\copyright~2022 IEEE. Personal use of this material is permitted.  Permission from IEEE must be obtained for all other uses, in any current or future media, including reprinting/republishing this material for advertising or promotional purposes, creating new collective works, for resale or redistribution to servers or lists, or reuse of any copyrighted component of this work in other works.
  \end{minipage}
}
\makeatother
\begin{document}

\title{Improving Multi-class Classifier Using Likelihood Ratio Estimation with Regularization}

\author{
\IEEEauthorblockN{Masato Kikuchi}
\IEEEauthorblockA{
\textit{Nagoya Institute of Technology}\\
Nagoya, Aichi, Japan \\
kikuchi@nitech.ac.jp
}
\and
\IEEEauthorblockN{Tadachika Ozono}
\IEEEauthorblockA{
\textit{Nagoya Institute of Technology}\\
Nagoya, Aichi, Japan \\
ozono@nitech.ac.jp
}
}
\maketitle

\begin{abstract}
The universal-set naive Bayes classifier (UNB)~\cite{Komiya:13}, defined using likelihood ratios (LRs), was proposed to address imbalanced classification problems.
However, the LR estimator used in the UNB overestimates LRs for low-frequency data, degrading the classification performance.
Our previous study~\cite{Kikuchi:19} proposed an effective LR estimator even for low-frequency data.
This estimator uses regularization to suppress the overestimation, but we did not consider imbalanced data.
In this paper, we integrated the estimator with the UNB.
Our experiments with imbalanced data showed that our proposed classifier effectively adjusts the classification scores according to the class balance using regularization parameters and improves the classification performance.
\end{abstract}


\section{Introduction}

Classifying data into appropriate categories is one of the most common problems in machine learning. 
Naive Bayes classifiers (NBs) are one of the most popular probabilistic classifiers.
They are often used due to following advantages: relatively good classification accuracy, high efficiency, and easy implementation.
Contrarily, NBs have various drawbacks.
Therefore, many researchers have attempted to improve them.
One of the drawbacks is that a classical NB cannot classify imbalanced data with sufficient accuracy.
In many real-world data, the proportions of classes to which instances belong are imbalanced.
In addition, some data have extremely imbalanced class proportions.
However, the classical NB is sparsely modeled for minority classes and misclassifies many instances as these classes.

One way to deal with imbalanced data is using complement classes.
This avoids uninformative modeling of classifiers and can improve classification performance.
As a classifier using complement classes, the universal-set naive Bayes classifier (UNB)~\cite{Komiya:13} was proposed.
It is defined as
\begin{align*}
& \widehat{c}(y) = \argmax_{c \in C} \frac{p(c)}{p(\bar{c})} \prod_{k=1}^{n} r(w_k, c), \\
& r(w_k, c) = \frac{p(w_k \mid c)}{p(w_k \mid \bar{c})},
\end{align*}
where $y=\langle w_1, w_2, \ldots, w_k, \ldots, w_n \rangle$ is an instance.
$w_k$ is the $k$-th attribute value of $y$ and is a countable discrete value such as a letter or word.
$c \in C$ is a class, and $\bar{c}$ is a complement class for $c$.
$\widehat{c}(y)$ is the class to which the classifier predicts $y$ belongs.
$r(w_k, c)$ is the likelihood ratio (LR).
The UNB handles imbalanced data.
However, this classifier cannot classify imbalanced data with sufficient accuracy since the LR estimator used in this classifier has problems.

\begin{table}[tb]
\caption{Estimation examples of LRs. We set $\lambda$ to $10^{-5}$.}
\label{tab:LR_example}
\centering
	\begin{tabular}{ c  r  r  r  r  r  r } \hline
		 \multicolumn{1}{ c }{\multirow{2}{*}{$x$}} & \multicolumn{4}{ c }{Observed Frequencies} & \multicolumn{1}{ c }{\multirow{2}{*}{$r_{\text{MLE}}(x)$}} & \multicolumn{1}{ c }{\multirow{2}{*}{$\widehat{r}(x)$}} \\ \cline{2-5}
		& \multicolumn{1}{ c }{$n_{\rm de}$} & \multicolumn{1}{ c }{$f_{\text{de}}(x)$} & \multicolumn{1}{ c }{$n_{\rm nu}$} & \multicolumn{1}{ c }{$f_{\text{nu}}(x)$} & &  \\ \hline
		$x_{\text{a}}$ & $10^{7}$ & 2,000 & $10^{4}$ & 100 & 50 & 47.6 \\
		$x_{\text{b}}$ & $10^{7}$ & 20 & $10^{4}$ & 1 & 50 & 8.3 \\
		$x_{\text{c}}$ & $10^{7}$ & 20 & $10^{4}$ & 2 & 100 & 16.7 \\ \hline
	\end{tabular}
\end{table}
The UNB estimates the two probabilities $p(w_k \mid c)$ and $p(w_k \mid \bar{c})$ as the relative frequencies and estimates $r(w_k, c)$ by taking their ratio.
However, our previous study~\cite{Kikuchi:19} suggested that this estimator overestimates LRs for infrequent events.
We explain this problem using LR estimation examples.
Let us define the LR $r(x)$ as
\begin{align*}
r(x) = \frac{p_{\rm nu}(x)}{p_{\rm de}(x)}.
\end{align*}
Let us define the estimator $r_{\rm MLE}(x)$ used by the UNB as
\begin{align*}
r_{\rm MLE}(x) = \frac{\widehat{p}_{\rm nu}(x)}{\widehat{p}_{\rm de}(x)}, \quad
\widehat{p}_{\rm *}(x) = \frac{f_{\rm *}(x)}{n_{\rm *}}.
\end{align*}
$* \in \{{\rm de, nu}\}$, where ``${\rm de}$'' and ``${\rm nu}$'' are indices representing the denominator and numerator of $r(x)$, respectively.
$f_{\rm *}(x)$ is the frequency of $x$ sampled from a probability distribution with density $p_{\rm *}(x)$, and $n_{\rm *}=\sum_x f_{\rm *}(x)$.
Suppose that the frequencies shown in Table~\ref{tab:LR_example} are given for events $x_{\rm a}$, $x_{\rm b}$, and $x_{\rm c}$.
Focusing on $x_{\rm a}$ and $x_{\rm b}$, although the frequencies $f_{\rm *}(x_{\rm a})$ and $f_{\rm *}(x_{\rm b})$ are different, their estimates are large (both 50).
However, the occurrence of $x_{\rm b}$ may be a coincidence since $f_{\rm de}(x_{\rm b})$ is the only one.
Minority classes rarely contain attribute values $w_k$s, and contained values are infrequent.
Therefore, overestimating LRs of infrequent $w_k$s results in unreasonably high classification scores for minority classes.
Thus, many instances are misclassified as minority classes.
Although the frequency difference between $f_{\rm nu}(x_{\rm b})$ and $f_{\rm nu}(x_{\rm c})$ is one, their estimates are very different ($50$ and $100$, respectively).
This fact means that the classification scores of the UNB are unstable for low frequencies.
This effect also contributes to classification performance degradation.

In our previous study~\cite{Kikuchi:19}, we proposed an LR estimator that can mitigate the above problem.
This estimator is defined as
\begin{align}
\widehat{r}(x) = \left\{ \frac{f_{\rm de}(x)}{n_{\rm de}} + \lambda \right\}^{-1} \frac{f_{\rm nu}(x)}{n_{\rm nu}}. \nonumber
\end{align}
It is derived in an optimization framework for the least-squares problem.
$\lambda$ ($\ge 0$) is a regularization parameter introduced in the optimization.
It can avoid the overestimation of LRs depending on the observed frequencies.
As shown in Table~\ref{tab:LR_example}, $\widehat{r}(x_a)=47.6$ estimated from high frequencies is approximately $50$.
Whereas, $\widehat{r}(x_b)=8.3$ estimated from low frequencies is significantly lower than 50.
Moreover, $\widehat{r}(x_c)=16.7$ estimated from low frequencies is significantly lower than 100.
Therefore, $\widehat{r}(x)$ provides lower (conservative) and stable estimates for low-frequency events.

Therefore, we combine the conservative LR estimator with the UNB.
In addition, we prepare different regularization parameters for each class and vary them according to the class balance in the data.
We experiment with classifying imbalanced data.
The experimental results show that the UNB misclassifies many instances into minority classes.
Due to this, the classification performance is sometimes significantly lower than that of the classical NB.
Our classifier assigns large parameter values for minority classes to avoid increasing classification scores.
As a result, it maintains a sufficient classification performance, even if the training dataset contains an extreme minority class.

\section{Related Work}

Many researchers have proposed various methods for imbalanced data classification.
These methods can be broadly categorized into data-level approaches and model-level approaches.
The most famous data-level approaches are upsampling~\cite{He:08} and downsampling~\cite{Zhang:03}.
These sampling methods measure the distance between the input data.
They artificially increase or decrease the sample size using the distance.
The data-level approaches can be applied to any classifier.
However, their utility is limited for data containing discrete values, such as text data. 
This is because it is difficult to define the distance between the data.
As a model-level approach, NBs using complement classes were proposed~\cite{Rennie:03,Komiya:13}.\footnote{For comparisons of these NBs with our classifier, see Appendix.}
Our classifier using complement classes belongs to the model-level approach.
NBs using cost-sensitive learning~\cite{Elkan:01} were also proposed in~\cite{Fang:12,Xiong:21}.
These NBs introduce different ``costs'' for misclassification of each class and conduct training and classification to minimize the expected cost.
Although setting optimal costs requires expert knowledge, it may be easier to fine-tune the classification performance for each class.
It is possible if the regularization parameters are adjusted using such costs for our classifier.

The indirect LR estimation approach, which estimates the probability distributions and takes their ratio, has a considerable estimation error~\cite{Hardle:04}.
For this reason, several direct LR estimation methods were proposed to estimate LR without estimating the probability distributions~\cite{Bickel:07,Sugiyama:08,Kanamori:09}.
However, these methods estimate LRs defined in continuous sample spaces and assume continuous values as sampled elements.
Therefore, we modified the basis functions used in the least-squares-based approach called unconstrained least-squares importance fitting (uLSIF)~\cite{Kanamori:09}.
It can estimate LRs defined in discrete sample spaces~\cite{Kikuchi:19}.
In addition, we proposed an LR estimation method for unobserved N-grams in a training dataset by estimating LRs of individual components of an N-gram and taking their product~\cite{Kikuchi:21}.
Then, we showed the effectiveness of this estimation method in binary classification.
This method is inspired by NBs and uses the results of our previous study~\cite{Kikuchi:19} for LR estimation.
In this paper, we apply this method to multi-class classification.

\section{Preliminaries}

We describe the UNB~\cite{Komiya:13} and conservative LR estimator~\cite{Kikuchi:19}, which are necessary to introduce our classifier.

\subsection{Universal-set Naive Bayes Classifier}
\label{sec:UNB}

Let $y= \langle w_1, w_2, \ldots, w_n \rangle$ be an instance consisting of discrete values $w_k$s ($k=1, 2, \ldots, n$), such as letters and words.
Let $c$ and $\bar{c}$ be a class and complement class, respectively.
The UNB~\cite{Komiya:13} is the probabilistic classifier using complement classes.
For $p(c \mid y)$ and $p(\bar{c} \mid y)$,
\begin{align*}
p(c \mid y) + p(\bar{c} \mid y) = 1
\end{align*}
holds.
By applying Bayes' theorem to $p(c \mid y)$ and $p(\bar{c} \mid y)$, respectively, we can transform the above equation into
\begin{align*}
\frac{p(y \mid c)p(c)}{p(y)} + \frac{p(y \mid \bar{c})p(\bar{c})}{p(y)} = 1.
\end{align*}
Solving this equation for $p(y)$, we obtain
\begin{align*}
p(y) = p(y \mid c)p(c) + p(y \mid \bar{c})p(\bar{c}).
\end{align*}
Substituting this into $p(c \mid y)$, we obtain
\begin{align*}
p(c \mid y) &=  \frac{p(y \mid c)p(c)}{p(y)} \\
&= \frac{p(y \mid c)p(c)}{p(y \mid c)p(c) + p(y \mid \bar{c})p(\bar{c})} \\
&= \frac{1}{1 + \frac{1}{\alpha}},
\end{align*}
where $\alpha$ is
\begin{align*}
\alpha = \frac{p(y \mid c)p(c)}{p(y \mid \bar{c})p(\bar{c})} = \frac{p(w_1, w_2, \ldots, w_n \mid c)p(c)}{p(w_1, w_2, \ldots, w_n \mid \bar{c})p(\bar{c})}.
\end{align*}
$p(c \mid y)$ is maximized when $\alpha$ is maximized.
We also assume conditional independence between $w_k$ and $w_{k'}$ ($k \ne k'$) under $c$ and $\bar{c}$.
By this assumption, the UNB is formulated as
\begin{align}
\label{eq:UNB}
\widehat{c}(y) = \argmax_{c \in C} \frac{p(c)}{p(\bar{c})} \prod_{k=1}^{n} r_{\rm MLE}(w_k, c),
\end{align}
where the LR $r(w_k, c)$ is estimated as
\begin{align}
\label{eq:LR_of_UNB}
r_{\rm MLE}(w_k, c) = \frac{\widehat{p}(w_k \mid c)}{\widehat{p}(w_k \mid \bar{c})}.
\end{align}
As shown in Eq. (\ref{eq:LR_of_UNB}), this classifier models the probability estimators $\widehat{p}(w_k \mid c)$ and $\widehat{p}(w_k \mid \bar{c})$ as the relative frequencies and takes their ratio to obtain $r_{\rm MLE}(w_k, c)$.
This estimation procedure is general and simple, but it often overestimates LRs based on low frequencies.
In class-imbalanced classification problems, the frequencies of $w_k$s vary greatly among classes.
The frequencies in minority classes are significantly lower than those in other classes.
Here, LRs for minority classes become unreasonably high compared to those for the others, missclassifying many instances as minority classes.
Therefore, we require a method to suppress the overestimation of LRs.

\subsection{Conservative Estimator for Likelihood Ratios}

We describe a problem setup for LR estimation.
Let $D \subset \mathbb{U}$ be a set of discrete values $x$s that a dataset contains.
Let $\mathbb{U}$ be a set of all $v$ types of values that can exist, which is also called a finite alphabet in information theory.
Suppose we obtain two samples from the probability distributions with density $p_{\rm de}(x)$ and $p_{\rm nu}(x)$, respectively:
\begin{align*}
\{x_i^{\rm de}\}_{i=1}^{n_{\rm de}} \overset{\rm i.i.d.}{\sim} p_{\rm de}(x)\ {\rm and}\ \{x_j^{\rm nu}\}_{j=1}^{n_{\rm nu}} \overset{\rm i.i.d.}{\sim} p_{\rm nu}(x),
\end{align*}
where $x$ refers to a linguistic element such as a letter and word.
Following previous studies, we assume that
\begin{align*}
p_{\rm de}(x) > 0 \quad {\rm for\ all\ }x \in D
\end{align*}
is satisfied.
This assumption allows us to define LRs for all $x$s.
In this section, we estimate the LR 
\begin{align*}
r(x) = \frac{p_{\rm nu}(x)}{p_{\rm de}(x)}
\end{align*}
directly using the two samples $\{x_i^{\rm de}\}_{i=1}^{n_{\rm de}}$ and $\{x_j^{\rm nu}\}_{j=1}^{n_{\rm nu}}$, without estimating the probability distributions.

ULSIF~\cite{Kanamori:09} is the direct LR estimation approach based on a least-squares fitting, defines the estimation model $\widehat{r}(x)$ as
\begin{align}
\label{eq:estimation_model}
\widehat{r}(x) = \sum_{l=1}^b \beta_l \varphi_l(x).
\end{align}
Here $\boldsymbol{\beta} = (\beta_1, \beta_2, \ldots, \beta_b)^{\rm T}$ are the parameters learned from the samples, and $\{\varphi_l\}_{l=1}^b$ are non-negative basis functions.
The original uLSIF deals with LRs defined on continuous spaces.
Therefore, to use the structure of the spaces, it uses basis functions with Gaussian kernels.
However, this paper deals with discrete values such as letters and words, and LRs are also defined in discrete spaces.
That is, Gaussian kernels are ineffective.
Hence, we substitute the basis functions $\{\delta_l\}_{l=1}^v$:
\begin{align}
\label{eq:basis_function}
\delta_l(x) = \left\{
\begin{array}{ll}
1 & {\rm if\ } x = x_{(l)}, \\
0 & {\rm otherwise},
\end{array}
\right.
\end{align}
which were proposed in our previous study~\cite{Kikuchi:19}.
These functions are defined for each value type. 
$l$ is an index that specifies the value type.
$x_{(l)}$ indicates the $l$-th value of all the $v$ types of values.
Using $\{\delta_l\}_{l=1}^v$, we can derive simple estimators for LRs on discrete spaces.
Here, $b$ is replaced by $v$.
We substitute Eq. (\ref{eq:basis_function}) into Eq. (\ref{eq:estimation_model}) to derive $\boldsymbol{\beta}$ that minimizes the squared error between $\widehat{r}(x)$ and the true LR $r(x)$.\footnote{The original reference~\cite{Kikuchi:19}, in which the derivation of $\boldsymbol{\beta}$ is described, is written in Japanese.
Thus, see~\cite{Kanamori:09} and Section III A of~\cite{Kikuchi:21} for the derivation.}
The resulting estimator $\widehat{r}\left( x_{(m)} \right)$ for $x_{(m)}\ (m=1, 2, \ldots, v)$ is
\begin{align}
\widehat{r}\left( x_{(m)} \right) = \left\{ \frac{f_{\rm de}\left( x_{(m)} \right)}{n_{\rm de}} + \lambda \right\}^{-1} \frac{f_{\rm nu}\left( x_{(m)} \right)}{n_{\rm nu}}, \nonumber
\end{align}
where $f_{\rm *}\left( x_{(m)} \right)$ is the frequency of $x_{(m)}$ sampled from the probability distribution with density $p_{\rm *}\left( x_{(m)} \right), * \in \{{\rm de, nu}\}$, and $n_{\rm *} = \sum_x f_{\rm *}(x)$.
In the above equation, the regularization parameter $\lambda\ (\ge 0)$ gives a lower (conservative) estimate depending on the frequencies.

To ensure that the product of LRs does not become zero or infinity, we use the LR estimator
\begin{align}
\label{eq:conservative_LR}
\widetilde{r}\left( x_{(m)} \right) &= \left\{ \frac{f_{\rm de}\left( x_{(m)} \right)+1}{n_{\rm de}+2} + \lambda \right\}^{-1} \frac{f_{\rm nu}\left( x_{(m)} \right)+1}{n_{\rm nu}+2}
\end{align}
in our classifier.
This LR estimator uses the correction frequencies.
The probability estimator $\frac{f_{*} \left( x_{(m)} \right) + 1}{n_{*} + 2}$ is equal to the expected a posteriori estimator when the probability distribution is a Bernoulli distribution of whether $x_{(m)}$ occurs or not, and the prior is a uniform distribution in the interval $[0,1]$.

\section{Our Proposed Classifier}
\label{sec:our_classifier}

The UNB estimates the probabilities by relative frequencies and takes their ratio to obtain the estimator $r_{\rm MLE}(w_k, c)$ in Eq. (\ref{eq:LR_of_UNB}).
However, $r_{\rm MLE}(w_k, c)$ overestimates LRs based on low frequencies.
As mentioned in Section~\ref{sec:UNB}, using this estimator causes many instances to be misclassified as minority classes.
To prevent this, we use Eq. (\ref{eq:conservative_LR}) for LR estimation.
Therefore, our classifier is formulated as
\begin{align}
\label{eq:our_classifier}
\widehat{c}(y) = \argmax_{c \in C} \frac{p(c)}{p(\bar{c})} \prod_{k=1}^{n} \widetilde{r}(w_k, \lambda_c).
\end{align}
It has the regularization parameters $\boldsymbol{\lambda}=\{ \lambda_c \mid c \in C \}$ for each class.
The parameters adjust the classification scores for each class according to the class balance.
Thus, they prevent many instances from being misclassified as minority classes.
For this purpose, $\widetilde{r}(w_k, \lambda_c)$ is given the parameter $\lambda_c$ as an argument for the class $c$, instead of $c$ itself.

This section describes how to determine $\boldsymbol{\lambda}$.
Let $\Theta={\{ 10^{-9}, 10^{-8}, \ldots, 10^{-1} \}}$ be the candidate values that $\lambda_c$ takes.
We find the combination $\boldsymbol{\lambda}^{'}=\{ \lambda_c^{'} \mid c \in C \}$ that maximizes an evaluation function $J$ based on the classification results.
In an $\rm M$-class classification, this optimization problem is formulated as 
\begin{align*}
\boldsymbol{\lambda}^{'} = \argmax_{\boldsymbol{\lambda} \in \Theta^{\rm M}} J({\rm MCC},\ \boldsymbol{\lambda}),
\end{align*}
where MCC is a multi-class classifier with $\boldsymbol{\lambda}$ as parameters.
Here, it is our classifier shown in Eq. (\ref{eq:our_classifier}).
Since we deal with imbalanced data, we set the function $J$ as a macro-averaged F1 score.
This score is low if MCC misclassifies many instances as a minority class.
Therefore, we use the regularization parameters to maximize the macro-averaged F1 score.
It will yield a good classification performance for each class on average.

Note that the total number of possible combinations of parameter values is $9^{\rm M}$.
When the number of classes $\rm M$ is large, it is difficult to search all combinations.
To obtain approximate optimal values, we use the Differential Evolution algorithm (DE).
It is a well-known optimization algorithm.\footnote{We also can use other optimization algorithms such as the greedy algorithm and local search algorithm.}
Based on~\cite{Wu:11}, we set the parameters for the DE as follows:
\begin{itemize}
    \item The maximum evolutionary generation ${\rm MaxGen}=50$;
    \item The population size ${\rm NP}=30$;
    \item The differential weight ${\rm F}=0.8$;
    \item The crossover probability ${\rm CR}=0.6$.
\end{itemize}
Note that we set ${\rm MaxGen}$ to 50 for efficient computation.
We regard a fitness function, which is the objective function of the optimization, as the macro-averaged F1 score.
The parameter values $\boldsymbol{\lambda}^{'}$ that maximize the score are regarded as the optimal values.
We prepare a validation dataset, consider it an evaluation dataset, and solve classification problems to calculate the score.
We explain experimental data in Section~\ref{sec:data_and_conditions}.

\section{Experiments}

We classify the occurrence contexts of named entities (NEs; proper nouns such as location names and personal names in a corpus) into six or seven classes and clarify the effectiveness of our classifier.
We deal with seven NEs: LOCATION, ORGANIZATION (ORG.), DATE, MONEY, PERSON, PERCENT, and TIME.
The contexts are the word 10-grams on the left of NEs. 
We have three reasons for conducting the experiments.
First, while linguistic elements are rich in variety, they are often infrequent.
This property allows us to validate the effectiveness of regularization parameters in preventing the overestimation of LRs.
Second, the NEs vary significantly in their ease of occurrence.
As discussed in Section~\ref{sec:data_and_conditions}, the context of ORG. occurs 91,892 times, while the context of TIME occurs only 840 times.
Therefore, experimental data are highly imbalanced and suitable for validating the effectiveness of our classifier.
Third, the contexts are uniquely determined, allowing quantitative evaluation for classifiers.

\subsection{Experimental Datasets}
\label{sec:data_and_conditions}

\begin{figure}[tb]
  \centering
  \includegraphics[keepaspectratio, scale=0.45]{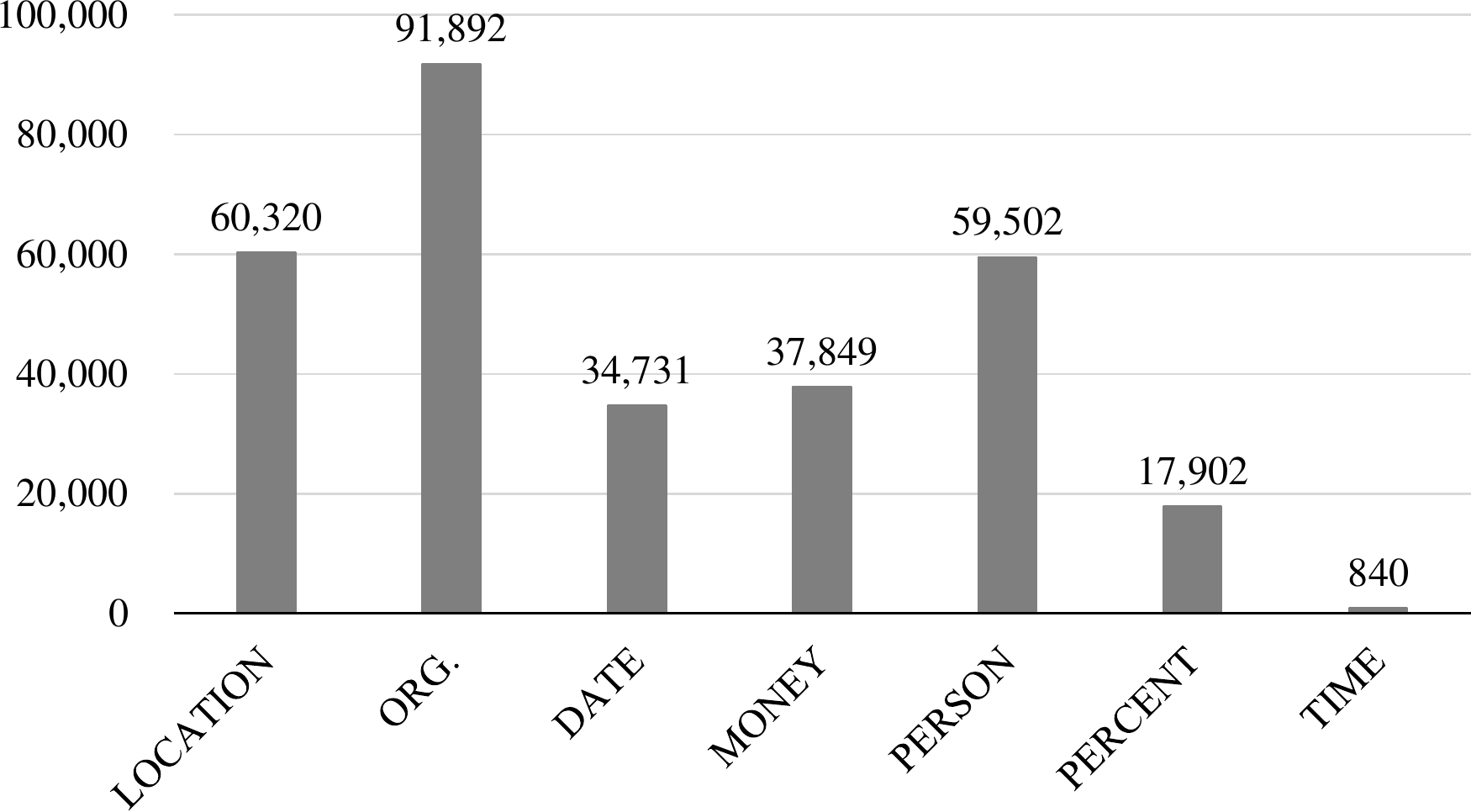}
  \caption{Instances contained in 10,000 training articles.}
  \label{fig:frequeicies}
\end{figure}
We created our experimental datasets based on the 1987 edition of the Wall Street Journal Corpus.\footnote{\url{https://catalog.ldc.upenn.edu/LDC2000T43}}
First, we randomly selected 12,000 articles from the corpus.
We divided 10,000, 1,000, and 1,000 articles into training, validation, and evaluation articles, respectively.
We then assigned NE tags to the articles using the Stanford named entity recognizer (Stanford NER)\footnote{\url{https://nlp.stanford.edu/software/CRF-NER.html}}~\cite{Finkel:05} and extracted the occurrence contexts (instances).
Fig.~\ref{fig:frequeicies} shows the instances contained in the training articles.
As shown in this figure, the context of TIME is extremely infrequent compared to the other contexts.
We use the instance sets extracted from the training, validation, and evaluation articles as the training, validation, and evaluation datasets.

\subsection{Experimental Procedure}
\label{sec:experimental_procedure}

We experiment with the following procedure.
We count frequencies to calculate the classification scores from the training dataset. 
For our classifier, we tune the regularization parameters $\lambda_c$s using the validation dataset.
We perform classification on all instances in the evaluation dataset.
To determine the performance of each classifier, we calculate a macro-averaged recall, precision, F1 score, and micro-averaged accuracy.

\subsection{Comparison Classifiers}

We compare the NB, UNB, and our classifier.

\noindent
\textbf{NB:}
We use the classical NB as a baseline.
We use Laplace smoothing to estimate $p(w_k \mid c)$.

\noindent
\textbf{UNB:}
The UNB is defined in Eq. (\ref{eq:UNB}) in Section \ref{sec:UNB}.
We correct the relative frequencies of $w_k$ by adding two to the denominator and one to the numerator
.\footnote{The smoothing techniques are commonly used to correct probability estimators.
However, if we use the smoothed estimates for LR estimation, resulting LR estimates before and after the correction may differ significantly, which causes overestimation.
Therefore, We slightly correct the relative frequencies.}
The LR estimator of the UNB is equal to the estimator with $\lambda=0$ in Eq. (\ref{eq:conservative_LR}).

\noindent
\textbf{Our classifier:}
Our classifier is defined in Eq. (\ref{eq:our_classifier}) in Section~\ref{sec:our_classifier}.
We use the DE to search the regularization parameters $\lambda_c$s and set different optimal values $\lambda_c^{'}$s for each class.

\subsection{Experimental Results}

\begin{table}[tb]
\centering
\caption{Optimal values $\lambda_c^{'}$ of regularization parameters obtained by the DE (left: the experiment using all classes, right: the experiment using six classes excluding TIME).}
\label{tab:regularization_parameters1}
\begin{tabular}{lr|lr} \hline
\multicolumn{1}{c}{Class} & \multicolumn{1}{c}{$\lambda_c^{'}$} & \multicolumn{1}{|c}{Class} & \multicolumn{1}{c}{$\lambda_c^{'}$} \\ \hline
\vspace{-2.4mm} & & & \\
LOCATION & $10^{-6}$ & LOCATION & $10^{-9}$ \\
ORG. & $10^{-9}$ & ORG. & $10^{-9}$ \\
DATE & $10^{-5}$ & DATE & $10^{-5}$ \\
MONEY & $10^{-5}$ & MONEY & $10^{-5}$ \\
PERSON & $10^{-5}$ & PERSON & $10^{-5}$ \\
PERCENT & $10^{-4}$ & PERCENT & $10^{-5}$ \\
TIME & $10^{-3}$ \\
\hline
\vspace{-3.5mm}
\end{tabular}
\end{table}
\begin{figure}[tb]
  \centering
  \includegraphics[keepaspectratio, scale=0.4]{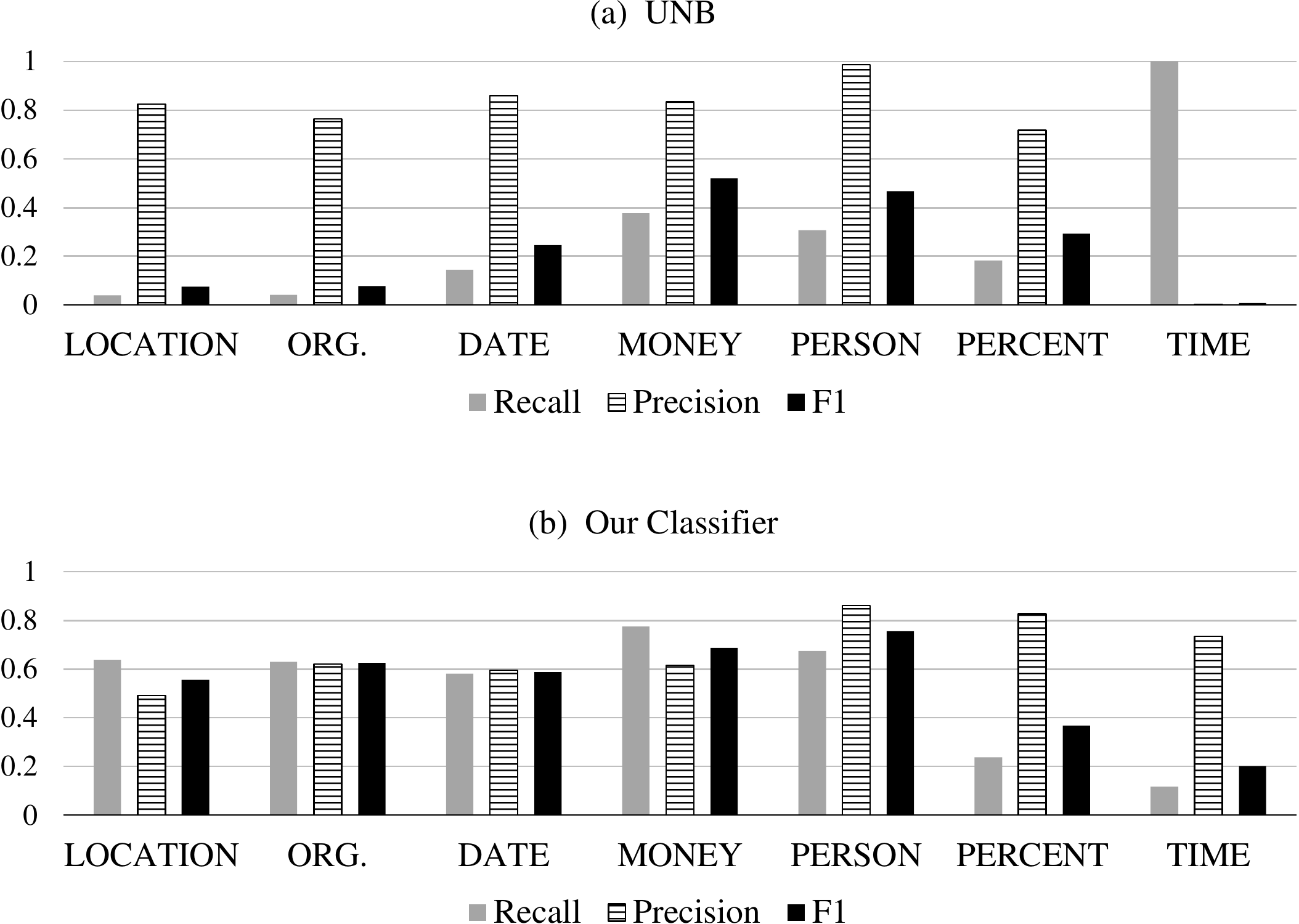}
  \caption{Recall, precision, and F1 scores for each class of the UNB and our classifier (all classes).}
  \label{fig:performance_class1}
\end{figure}
\begin{table}[tb]
\centering
\caption{Classification results using all classes.
The maximum values of the evaluation indicators are written in bold.}
\label{tab:results2-1}
\begin{tabular}{lrrrr} \hline
\multicolumn{1}{ c }{\multirow{2}{*}{Classifier}} & \multicolumn{3}{c}{Macro} & \multicolumn{1}{ c }{Micro} \\
& \multicolumn{1}{ c }{R} & \multicolumn{1}{ c }{P} & \multicolumn{1}{ c }{F1} & \multicolumn{1}{ c }{A} \\ \hline
NB & {\bf 0.546} & 0.510 & 0.483 & 0.529 \\
UNB & 0.299 & {\bf 0.713} & 0.241 & 0.160 \\
Ours & 0.522 & 0.678 & {\bf 0.540} & {\bf 0.626} \\ \hline
\end{tabular}
\end{table}
This section describes the classification results using the training data shown in Fig.~\ref{fig:frequeicies} in Section~\ref{sec:data_and_conditions}.
Table~\ref{tab:regularization_parameters1} left shows the optimal values $\lambda_c^{'}$s of the regularization parameters used in our classifier.
This table and Fig.~\ref{fig:frequeicies} show that the majority and minority classes are assigned smaller and larger parameter values, respectively.
This tendency indicates that our classifier adjusts the regularization parameters according to the class sizes.
As shown in Table~\ref{tab:results2-1}, the F1 score and accuracy of the UNB are low, which does not allow for adequate classification.
Contrarily, our classifier shows the best F1 score and accuracy, suggesting the effectiveness of regularization parameters.
For the detailed analysis of the performance differences between the UNB and our classifier, we show the recall, precision, and F1 scores for each class in Fig.~\ref{fig:performance_class1}.
This figure shows that the UNB has a high recall but low precision and F1 score for TIME.
For classes except TIME, only precision is high.
This result indicates that the UNB misclassifies many instances to TIME.
On the other hand, our classifier has low F1 scores for PERCENT and TIME but improved F1 scores for the others.
Additionally, it prevents misclassification into the minority classes.

\begin{table}[tb]
\centering
\caption{Classification results using six classes excluding TIME.
The maximum values of the evaluation indicators are written in bold.}
\label{tab:results2-2}
\begin{tabular}{lrrrr} \hline
\multicolumn{1}{ c }{\multirow{2}{*}{Classifier}} & \multicolumn{3}{c}{Macro} & \multicolumn{1}{ c }{Micro} \\
& \multicolumn{1}{ c }{R} & \multicolumn{1}{ c }{P} & \multicolumn{1}{ c }{F1} & \multicolumn{1}{ c }{A} \\ \hline
NB & 0.609 & 0.574 & 0.571 & 0.582 \\
UNB & 0.610 & 0.578 & 0.564 & 0.568 \\
Ours & {\bf 0.624} & {\bf 0.624} & {\bf 0.610} & {\bf 0.617} \\ \hline
\end{tabular}
\end{table}
\begin{figure}[tb]
  \centering
  \includegraphics[keepaspectratio, scale=0.4]{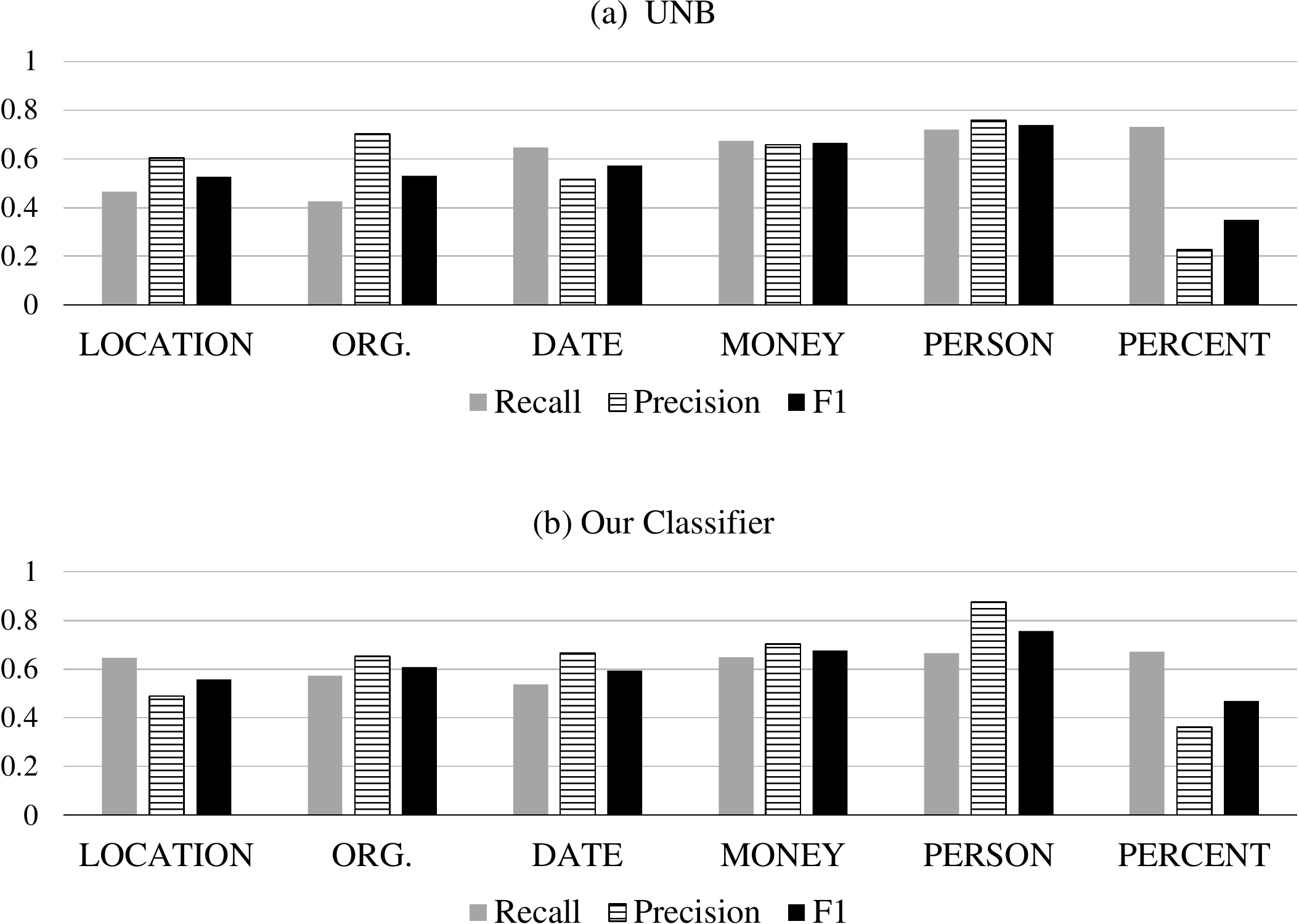}
  \caption{Recall, precision, and F1 scores for each class of the UNB and our classifier (excluding TIME).}
  \label{fig:performance_class2}
\end{figure}
TIME is an extreme minority class.
Therefore, we removed the contexts of TIME from the training, validation, and evaluation datasets and then performed a six-class classification using each classifier.
Table~\ref{tab:regularization_parameters1} right shows the optimal $\lambda_c^{'}$s of the regularization parameters.
Although the difference among optimal values is smaller than in that of all classes, there exists a tendency to assign smaller and larger values to the majority and minority classes, respectively.
As shown in Table~\ref{tab:results2-2}, the performance of the UNB increases significantly when TIME is excluded from the classification.
However, the F1 score and accuracy are lower than those for the NB, suggesting that the UNB needs improvement.
Fig.~\ref{fig:performance_class2} shows the recall, precision, and F1 scores for each class of the UNB and our classifier.
As shown in this figure, there is no significant performance difference between the two classifiers.
We compare Figs.~\ref{fig:performance_class1}(b) and~\ref{fig:performance_class2}(b) for our classifier.
For PERCENT, the recall and precision are varied widely, however F1 scores are approximately equal.
In other classes, there is no significant fluctuation in any indicators.
This result suggests that our classifier has the advantage of significantly preserving the classification performance of majority classes, even if some of the classes are extreme minority classes.

\begin{table*}[tb]
\centering
\caption{Recall, precision, and F1 scores for each class.}
\label{tab:results_all}
\begin{tabular}{l r rrr rrr r rrr r rrr} \hline
\multicolumn{1}{ c }{\multirow{2}{*}{Classifier}} & \multicolumn{3}{c}{LOCATION} & & \multicolumn{3}{c}{ORG.} & & \multicolumn{3}{c}{DATE} & & \multicolumn{3}{c}{MONEY} \\
& \multicolumn{1}{ c }{${\rm R}_c$} & \multicolumn{1}{ c }{${\rm P}_c$} & \multicolumn{1}{ c }{${\rm F1}_c$} & & \multicolumn{1}{ c }{${\rm R}_c$} & \multicolumn{1}{ c }{${\rm P}_c$} & \multicolumn{1}{ c }{${\rm F1}_c$} & & \multicolumn{1}{ c }{${\rm R}_c$} & \multicolumn{1}{ c }{${\rm P}_c$} & \multicolumn{1}{ c }{${\rm F1}_c$} & & \multicolumn{1}{ c }{${\rm R}_c$} & \multicolumn{1}{ c }{${\rm P}_c$} & \multicolumn{1}{ c }{${\rm F1}_c$} \\ \hline
NB & 0.416 & 0.601 & 0.491 & & 0.439 & 0.662 & 0.528 & & 0.579 & 0.526 & 0.551 & & 0.666 & 0.642 & 0.654 \\
CNB & 0.519 & 0.586 & $\bullet\ $0.550 & & 0.718 & 0.570 & $\clubsuit\ $0.636 & & 0.421 & 0.716 & 0.530 & & 0.726 & 0.604 & $\bullet\ $0.659 \\
CNB\ \ ($p(c)=1$)& 0.460 & 0.634 & 0.533 & & 0.451 & 0.687 & 0.545 & & 0.488 & 0.634 & 0.551 & & 0.703 & 0.605 & 0.650 \\
NNB & 0.489 & 0.616 & 0.545 & & 0.540 & 0.652 & 0.591 & & 0.491 & 0.643 & $\bullet\ $0.557 & & 0.727 & 0.601 & 0.658 \\
UNB & 0.040 & 0.824 & 0.077 & & 0.041 & 0.764 & 0.078 & & 0.144 & 0.860 & 0.247 & & 0.377 & 0.833 & 0.519 \\
Ours & 0.639 & 0.492 & $\clubsuit\ $0.556 & & 0.630 & 0.619 & $\bullet\ $0.625 & & 0.582 & 0.595 & $\clubsuit\ $0.588 & & 0.774 & 0.616 & $\clubsuit\ $0.686 \\
\hline
\multicolumn{1}{ c }{\multirow{2}{*}{Classifier}} & \multicolumn{3}{c}{PERSON} & & \multicolumn{3}{c}{PERCENT} & & \multicolumn{3}{c}{TIME} \\
& \multicolumn{1}{ c }{${\rm R}_c$} & \multicolumn{1}{ c }{${\rm P}_c$} & \multicolumn{1}{ c }{${\rm F1}_c$} & & \multicolumn{1}{ c }{${\rm R}_c$} & \multicolumn{1}{ c }{${\rm P}_c$} & \multicolumn{1}{ c }{${\rm F1}_c$} & & \multicolumn{1}{ c }{${\rm R}_c$} & \multicolumn{1}{ c }{${\rm P}_c$} & \multicolumn{1}{ c }{${\rm F1}_c$} \\ \cline{1-12}
NB & 0.637 & 0.798 & 0.708 & & 0.598 & 0.331 & $\clubsuit\ $0.427 & & 0.489 & 0.011 & $\bullet\ $0.021 \\
CNB & 0.749 & 0.692 & 0.720 & & 0.204 & 0.831 & 0.327 & & 0 & 0 & 0 \\
CNB\ \ ($p(c)=1$)& 0.727 & 0.734 & $\bullet\ $0.730 & & 0 & 0 & 0 & & 0.404 & 0.005 & 0.010 \\
NNB & 0.738 & 0.711 & 0.724 & & 0 & 0 & 0 & & 0.287 & 0.006 & 0.011 \\
UNB & 0.307 & 0.987 & 0.468 & & 0.183 & 0.717 & 0.292 & & 1 & 0.004 & 0.007 \\
Ours & 0.675 & 0.861 & $\clubsuit\ $0.757 & & 0.237 & 0.826 & $\bullet\ $0.369 & & 0.117 & 0.733 & $\clubsuit\ $0.202 \\ \cline{1-12}
\end{tabular}
\end{table*}

\section{Conclusion}

This paper proposed a new classifier by combining the conservative LR estimator with the UNB.
This classifier uses the regularization parameters $\lambda_c$s introduced in the LR estimation process to adjust the classification scores according to the class balance.
In the experiments using imbalanced data, the UNB overestimated LRs for minority classes.
Its performance was lower than that of the NB (the macro-averaged F1: 0.241).
Contrarily, our classifier suppressed the overestimation of LRs with $\lambda_c$s and achieved the best performance (the macro-averaged F1: 0.540).

However, as shown in Fig.~\ref{fig:performance_class1}, the F1 scores of the minority classes are low even for our classifier.
In actual classification tasks, it is often desirable to improve the classification accuracy for minority classes at the expense of accuracy for others.
Since our classifier can adjust the classification scores for each class, we can improve the performance of minority classes by changing the search process of $\lambda_c$s.
For example, using cost-sensitive learning, we can define high costs for classification failure of minority classes and search $\lambda_c$s to minimize the expected entire cost. 
The extension of the classifier for practical use, as described above, will be studied in our future works.

\section*{Acknowledgment}

This work was supported in part by JSPS KAKENHI Grant Numbers JP19K12266, JP22K18006.

\section*{Appendix}

We compare our classifier with the complement naive Bayes classifier (CNB)~\cite{Rennie:03} and negation naive Bayes classifier (NNB)~\cite{Komiya:13}, which use complement classes.
We conduct the experiments using the procedure described in Section~\ref{sec:experimental_procedure}.

\noindent
\textbf{CNB:}
The CNB, a well-known classifier using complement classes, is defined as
\begin{align*}
\widehat{c}(y) = \argmax_{c \in C} p(c) \prod_{k=1}^{n} \frac{1}{p(w_k \mid \bar{c})}.
\end{align*}
We use Laplace smoothing to estimate $p(w_k \mid \bar{c})$.
In practical use, $p(c)$ is ignored and only $p(w_k \mid \bar{c})$ is estimated.
Thus, we also include the CNB with $p(c)=1$ in the comparison.

\noindent
\textbf{NNB:}
The CNB is a heuristic classifier and cannot be derived from the posterior probability maximization formula.
The NNB, which is derived from the formula and uses the complement classes, is defined as
\begin{align*}
\widehat{c}(y) = \argmax_{c \in C} \frac{1}{1-p(c)} \prod_{k=1}^{n} \frac{1}{p(w_k \mid \bar{c})}.
\end{align*}
We use Laplace smoothing to estimate $p(w_k \mid \bar{c})$.

Table~\ref{tab:results_all} shows the classification results for each class.
We highlighted the largest and second-largest F1 scores by the symbols $\clubsuit$ and $\bullet$, respectively.
This table shows that our classifier achieves the largest F1 scores in five of the seven classes and second-largest F1 scores in the other two classes.
The CNB and NNB cannot classify any instances to PERCENT or TIME, the minority classes.
We set these classes' precision and F1 scores to zeros since they cannot be calculated.
The classifiers excluding the CNB and our classifier have low precision and F1 score for TIME.
They misclassify many instances to this class.
However, our classifier tends to classify only plausible instances to TIME.

\bibliography{references}

\begin{thebibliography}{10}

\bibitem{Komiya:13}
K.~Komiya, Y.~Ito, and Y.~Kotani.
\newblock New naive bayes methods using data from all classes.
\newblock {\em International Journal of Advanced Intelligence}, 5(1):1--12,
  2013.

\bibitem{Kikuchi:19}
M.~Kikuchi, K.~Kawakami, M.~Yoshida, and K.~Umemura.
\newblock Conservative direct estimation for likelihood ratios based on
  observed frequencies.
\newblock {\em IEICE Trans. Inf. \& Syst. (Japanese Edition)},
  J102-D(4):289--301, 2019.

\bibitem{He:08}
H.~He, Y.~Bai, E.~A. Garcia, and S.~Li.
\newblock {ADASYN}: {Adaptive} synthetic sampling approach for imbalanced
  learning.
\newblock In {\em Proc. IJCNN'08}, pages 1322--1328, 2008.

\bibitem{Zhang:03}
J.~Zhang and I.~Mani.
\newblock {kNN} approach to unbalanced data distributions: {A} case study
  involving information extraction.
\newblock In {\em Proc. the ICML'03 Workshop on Learning from Imbalanced
  Datasets}, pages 1--7, 2003.

\bibitem{Rennie:03}
J.~D.~M. Rennie, L.~Shih, J.~Teevan, and D.~R. Karger.
\newblock Tackling the poor assumptions of naive bayes text classifiers.
\newblock In {\em Proc. ICML'03}, pages 616--623, 2003.

\bibitem{Elkan:01}
C.~Elkan.
\newblock The foundations of cost-sensitive learning.
\newblock In {\em Proc. IJCAI'01}, pages 973--978, 2001.

\bibitem{Fang:12}
X.~Fang.
\newblock Inference-based naive bayes: {Turning} naive bayes cost-sensitive.
\newblock {\em IEEE Transactions on Knowledge and Data Engineering},
  25(10):2302--2313, 2012.

\bibitem{Xiong:21}
Y.~Xiong, M.~Ye, and C.~Wu.
\newblock Cancer classification with a cost-sensitive naive bayes stacking
  ensemble.
\newblock {\em Computational and Mathematical Methods in Medicine}, pages
  1--12, 2021.

\bibitem{Hardle:04}
W.~H{\"a}rdle, A.~Werwatz, M.~M{\"u}ller, and S.~Sperlich.
\newblock {\em Nonparametric and Semiparametric Models}.
\newblock Springer, 2004.

\bibitem{Bickel:07}
S.~Bickel, M.~Br{\"u}ckner, and T.~Scheffer.
\newblock Discriminative learning for differing training and test
  distributions.
\newblock In {\em Proc. ICML'07}, pages 81--88, 2007.

\bibitem{Sugiyama:08}
M.~Sugiyama, S.~Nakajima, H.~Kashima, P.~von B{\"u}nau, and M.~Kawanabe.
\newblock Direct importance estimation with model selection and its application
  to covariate shift adaptation.
\newblock In {\em Advances in NIPS}, pages 1433--1440, 2008.

\bibitem{Kanamori:09}
T.~Kanamori, S.~Hido, and M.~Sugiyama.
\newblock A least-squares approach to direct importance estimation.
\newblock {\em Journal of Machine Learning Research}, 10:1391--1445, July 2009.

\bibitem{Kikuchi:21}
M.~Kikuchi, M.~Yoshida, K.~Umemura, and T.~Ozono.
\newblock Feature selective likelihood ratio estimator for low- and
  zero-frequency {N-grams}.
\newblock In {\em Proc. ICAICTA'21}, pages 1--6, 2021.

\bibitem{Wu:11}
J.~Wu and Z.~Cai.
\newblock Attribute weighting via differential evolution algorithm for
  attribute weighted naive bayes ({WNB}).
\newblock {\em Journal of Computational Information Systems}, 7(5):1672--1679,
  2011.

\bibitem{Finkel:05}
J.~R. Finkel, T.~Grenager, and C.~Manning.
\newblock Incorporating non-local information into information extraction
  systems by {G}ibbs sampling.
\newblock In {\em Proc. ACL'05}, pages 363--370, 2005.

\end{thebibliography}
\bibliographystyle{unsrt}

\end{document}